%% file: conference_101719.tex
\documentclass[conference]{IEEEtran}
\IEEEoverridecommandlockouts
\usepackage{cite}
\usepackage{amsmath,amssymb,amsfonts}
\usepackage{algorithmic}
\usepackage{graphicx}
\usepackage{textcomp}
\usepackage{xcolor}
\usepackage{float} 
\usepackage{hhline}
\usepackage{boldline} 
\usepackage{multirow}
\usepackage[hyphens,spaces,obeyspaces]{url}
\usepackage{transparent}
\usepackage{tikz}
\usepackage{lipsum}
\newcommand\copyrighttext{%
  \transparent{0.4}
  \textnormal{\scriptsize \copyright 2022 IEEE 978-1-6654-7406-1/22/\$31.00\\DOI 10.1109/CogMI56440.2022.00025 \\
{\scriptsize A Multilingual Virtual Guide for Self-Attachment Technique Alicia Jiayun Law, Ruoyu Hu, Lisa Alazraki, Anandha Gopalan, Neophytos Polydorou, Abbas Edalat 2022 IEEE 4th International Conference on Cognitive Machine Intelligence (CogMI)}
}}

\newcommand\copyrightnotice{%
\begin{tikzpicture}[remember picture,overlay]
\node[anchor=south east,yshift=15pt] at (current page.south) {\parbox{\linewidth}{\copyrighttext}};
\end{tikzpicture}%
}

\def\BibTeX{{\rm B\kern-.05em{\sc i\kern-.025em b}\kern-.08em
    T\kern-.1667em\lower.7ex\hbox{E}\kern-.125emX}}
\begin{document}

\title{A Multilingual Virtual Guide for Self-Attachment Technique\\
}

\author{
\IEEEauthorblockN{Alicia Jiayun Law{$^{1,\dagger}$}, Ruoyu Hu{$^{1, 2, \dagger}$}\thanks{$\dagger$ Equal contribution.}, Lisa Alazraki{$^1$}, Anandha Gopalan{$^1$}, Neophytos Polydorou{$^{1,2}$}, Abbas Edalat{$^1$}}
\IEEEauthorblockA{$^1$\textit{Department of Computing, Imperial College London, UK}}
\IEEEauthorblockA{$^2$\textit{UKRI Centre for Doctoral Training in AI for Healthcare, Imperial College London, UK}}
Email : \{alicia.law15, ruoyu.hu18, lisa.alazraki20, a.gopalan, neophytos.polydorou19, a.edalat\}@imperial.ac.uk
}

\maketitle

\begin{abstract}
\input{sections/0-abstract}
\copyrightnotice
\end{abstract}


\section{Introduction}
\input{sections/I-intro}

\section{Background}
\input{sections/II-background.tex}

\section{Dataset}
\input{sections/III-dataset.tex}

\section{Implementation}
\input{sections/IV-implementation.tex}

\section{Non-Clinical Trial}
\input{sections/V-trial.tex}

\section{Discussion}
\input{sections/VI-discussion.tex}

\section{Acknowledgement}
\input{sections/VII-acknowledgement.tex}


\bibliographystyle{./IEEEtran}
\bibliography{refs}

\end{document}

%% file: sections/0-abstract.tex
In this work, we propose a computational framework that leverages existing out-of-language data to create a conversational agent for the delivery of Self-Attachment Technique (SAT) in Mandarin. Our framework does not require large-scale human translations, yet it achieves a comparable performance whilst also maintaining safety and reliability. We propose two different methods of augmenting available response data through empathetic rewriting. We evaluate our chatbot against a previous, English-only SAT chatbot through non-clinical human trials ($N=42$), each lasting five days, and quantitatively show that we are able to attain a comparable level of performance to the English SAT chatbot. We provide qualitative analysis on the limitations of our study and suggestions with the aim of guiding future improvements.

%% file: sections/I-intro.tex

According to the 2022 Global Burden of Disease study, mental disorders have been ranked amongst the top ten leading causes of burden\footnote{Burden is defined according to a disease's prevalence and harm \cite{GBD2022Burden}.} worldwide since 1990 \cite{GBD2022Burden}. With the onset of the COVID-19 pandemic, there has been significant negative impact on the mental health condition of the global population from a variety of environmental stimuli \cite{Pfefferbaum2020Mental}, with, for example, the effect in the UK most severe among the 18-34 demographic group but visible in all age demographics \cite{Daly2020Longitudinal}. Cases of patients suffering mental health issues associated with a range of negative emotions such as defeat, entrapment and loneliness increased significantly from pre-pandemic levels \cite{OConnor2021Mental}.

As such, provision of mental health support has become more imperative in addressing mental health concerns arising as a result of public health emergencies. Yet, there persists a ``mental health treatment gap", which describes the large disparity between the need for and availability of mental healthcare services \cite{GBD2022Burden}. This can be attributed to the following reasons: (i) stigma on mental health\cite{Shang2019Mental}, (ii) unaffordable treatment and (iii) limited and unequal distribution of mental healthcare resources \cite{Qin2020Understanding}. It is therefore desirable to incorporate and supplement existing methods with digital technologies and novel techniques.

The Self-Attachment Technique (SAT) is a self-administrable intervention introduced in~\cite{Edalat2015Introduction},~\cite{Edalat20171} and~\cite{Edalat2017}. In SAT, the user enacts both the role of the care-seeker, conceptualised as their childhood or emotional self and represented by the user's favourite childhood photo or VR avatar created from the photo, and that of the care-giver, conceptualised as their adult or thinking self. The adult self establishes an imaginative compassionate relation and then an affectional bonding with the childhood self using the photo or the avatar and their favourite jolly and love songs. Subsequently, for the bulk of the SAT intervention, the adult self re-parents the childhood self to emotional and social maturity by emulating the optimal parent-child interactions whenever the user experiences strong negative emotions, which are projected and externalised onto the childhood self. SAT has had promising results in its pilot study~\cite{edalat2022pilot}. 

Prior works have incorporated technologies into the delivery of SAT protocols, with the most recent producing a chatbot assistant \cite{Alazraki2021Satbot} aimed at guiding users proficient with practising SAT protocols through protocol recommendation. Conversational agents \cite{Ji2022Achieving} have significant potential in their application to psychotherapy \cite{Sun2021PsyQA, Rashkin2019Towards}, as recent advancements in the field of Natural Language Processing with large neural pretrained language models \cite{Devlin2018Bert, Radford2019Language} using a transformer-based architecture \cite{Vaswani2017Attention} have achieved state-of-the-art results in a variety of tasks that facilitate greater capability of human-computer interaction.

However, prior works are limited only to English, a situation emblematic of much of the recent progress in the application of machine learning models to Natural Language Processing \cite{Hu2020xtreme, Xue2021mt5}. Monolingual NLP for certain languages can encounter the problem of resource availability, where there is a lower volume of available task-specific data to train a model to the same level of performance as higher-resource languages such as English.

In this paper, we present a computational framework for the delivery of SAT protocols in a Mandarin setting in order to gauge the feasibility of deploying existing English psychotherapeutic intervention into non-English languages, with the aim to contribute to achieving equitable access to mental healthcare for non-English speaking communities in the future. We summarise our contributions as follows:

\begin{itemize}
    \item We introduce a translation pipeline, leveraging machine translation and post-editing to produce language-specific data from existing task-specific English data.
    \item We introduce transformer reinforcement learning via Proximal Policy Optimisation (PPO) to train an empathetic, fluent and accurate generation model to produce quality responses via empathetic rewriting.
    \item We introduce an alternate, supervised learning method for empathetic rewriting and provide quantitative comparison against the previous methods.
    \item We introduce a multilingual emotion recognition component, and apply knowledge distillation to reduce inference latency.
    \item We fully integrate the Mandarin version of the chatbot with previous \cite{Alazraki2021Satbot} English versions to deploy a fully bilingual application.
    \item We formally evaluate the chatbot performance in multiple non-clinical trials, and provide qualitative analysis aimed at guiding future work.
\end{itemize}

%% file: sections/II-background.tex
\subsection{Self-Attachment Technique}
\label{background:self_attachment}

Self-Attachment Technique (SAT) \cite{Edalat2015Introduction} is a new psychotherapeutic treatment informed by John Bowlby's Attachment Theory and has shown promise in early pilot studies \cite{edalat2022pilot}. It attributes affect dysregulation\footnote{Affect dysregulation is defined as the ``impaired ability to regulate and/or tolerate negative emotional states" \cite{Dvir2014Childhood}.} disorders to sub-optimal emotional attachments formed between an individual and their primary caregivers during their early childhood. For instance, individuals who experienced secure attachment (i.e., had available and responsive caregivers) in their childhood tend to exhibit stronger self-esteem and self-reliance, and hence healthier mental states as adults \cite{Edalat2015Introduction}.

SAT is comprised of 20 self-administered protocols aimed at developing new secure attachment. The protocols invite individuals to envisage their current self caring and attending to their inner childhood self. This stimulates optimal neural growth, allowing individuals to better navigate and regulate their negative emotions, thereby tackling mental disorders stemming from insecure attachment \cite{Edalat20171}. The aims of the 20 protocols can be collated into eight groups:

\begin{itemize}
    \item Compassion toward the childhood self.
    \item Affectional bonding with the childhood self; Vowing to care for the childhood self.
    \item Rebuilding the childhood self's emotional world; Loving the childhood self, zest for life; Bonding with Nature.
    \item Self-regulation of strong emotions; Reducing negative emotions.
    \item (Re)-learning to laugh and being playful.
    \item Learning to change perspective and laugh.
    \item Socialising the childhood self.
    \item Enhancing tolerance and resilience.
\end{itemize}

The previous English SAT chatbot \cite{Alazraki2021Satbot} deduces the user's emotional state from open conversation, yet allows the user to select a different emotion whenever they feel that the one inferred is inaccurate. The chatbot then pursues a series of questions depending on the user's emotion to further refine protocol recommendation based on the user's past experience and current state. Protocols deemed unsuitable or those with which the user had previous adverse reactions, particularly protocols aimed at tackling negative emotions, are eliminated from recommendation. Users are encouraged to select and practise a protocol from a list of recommendations. Afterwards, they are prompted to give feedback on changes to their emotional state and undertake further protocols that are selected based on the feedback.

\subsection{Empathy in Digital Psychotherapy}
\label{background:empathy}
According to psychotherapy research, an important component in the efficacy of psychotherapeutic interventions is the capability of the therapist to engage in an empathetic manner with the patient \cite{Elliot2018Therapist}. Similarly to prior works \cite{Alazraki2021Satbot, Sharma2020Computational}, we focus on Godfrey T. Barrett-Lennard's second phase of empathetic dialogue with the aim of producing empathetic responses demonstrating compassion towards the user.

Prior works on empathetic dialogue systems \cite{Sharma2021Empathic, Rashkin2019Towards, Sharma2020Computational} have highlighted the importance of empathy in digital mental health support. However, most prior works focus on English deployment, and at the time of writing, there is little open-domain, language-specific and task-specific data for open empathetic response generation in Mandarin.


\subsection{Related works}
\label{background:chatbots}

Applications of digital psychotherapeutic interventions, such as \textit{Cognitive-Behavioural Therapy} (CBT) \cite{Wilhelm2020cbt} remain largely monolingual, though with increasing monolingual adoption in non-English languages in recent years, such as French in the case of Lopez et al. \cite{Lopez2019French}. Works such as those by Bakker et al. \cite{Bakker2018cbt} and Weaver et al. \cite{Weaver2021cbt} present English-only digital platforms for CBT and cite the adaptation to more languages as a future research direction to increase impact.

Previous SAT chatbots \cite{Alazraki2021Satbot} make use of crowd-sourced English data for their responses, and a fixed conversation flow that determines the appropriate response type at each conversation step. Responses are retrieved from a pool of possible responses ranked by a weighted metric combining sentence fluency, novelty to previous conversation, and perceived empathy. We aim to leverage the existing task-specific data produced by \cite{Alazraki2021Satbot} through translation.

Several approaches exist to facilitate multilinguality in chatbot response generation, though we focus broadly on two translation approaches:

\begin{itemize}
    \item \textbf{Inference-time} translations \cite{Ralston2019Voice, Lin2021XPersona}, wherein the semantics of the output utterance is determined prior to performing translation on a selected response. This has a relatively low data footprint, and changes to the translation system can be deployed immediately. However, it requires higher computational demand at inference-time. Translation mechanisms can be embedded within the response generation step such as in Gra\c ca et al. \cite{Graca2020Maia} and Dimitra et al. \cite{Dimitra2022Translation} to provide customer support and public administration functionalities respectively. Ralston et al. \cite{Ralston2019Voice} apply this approach to the provision of mental health support to students by wrapping conversation logic within source-target and target-source translation steps using multiple external APIs. Lin et al. \cite{Lin2021XPersona} identify the high cost of this approach, as well as the susceptibility to noisy data at inference-time, and propose learning language-agnostic representations to allow for better multilingual adaptation through training on translated data.
    \item \textbf{Pre-computed} \cite{Nieminen2022Coproducing} translations, wherein the responses are saved and retrieved at the relevant stages of the conversation. This approach allows for all the responses to be known prior to the retrieval step, allowing for finer control over the responses presented to the user. Due to the sensitive nature of conversations in mental health contexts, this feature is inherently beneficial for ensuring the safety of the produced responses. However, in order to deploy to multiple languages, this approach requires the creation of language files for each supported language, as seen in the mental health support chatbot from Nieminen et al. \cite{Nieminen2022Coproducing}. \textit{Retrieval-based} methods are nonetheless commonly used in conversational agents for mental health in a monolingual context, such as in Alazraki et al. \cite{Alazraki2021Satbot}, where a retrieval model allows the English SAT chatbot to guide users through carrying out self-attachment therapy protocols, and in Vaira et al. \cite{Vaira2018Mama}, where a similar model is used to provide support to new mothers.
\end{itemize}

For the purpose of this paper, we focus on pre-computing responses in order to ensure the safety and reliability of the conversation provided to the user, as well as to prevent translation errors that may negatively impact user experience.

\subsection{Patient Safety}

The SAT chatbot is a mental health application targeted at patients suffering from mental health conditions. In the interest of their safety, we take the following measures:
\begin{enumerate}
    \item \textit{Safe \& Non-Toxic Chatbot Conversations} \\
    Empathetic rewritings are produced from generative language models trained using a controlled dataset that has been vetted for safety. Utterances are also scored for empathy via an empathy classifier (the scores assigned are discrete labels) and only those that have attained a high empathy score are selected. Finally, as an additional precaution, all utterances are manually vetted for toxic content before being included in the dataset.
    
    \item \textit{Terminating Therapy} \\
    SAT Protocols involve users interacting with their childhood self, which can inadvertently trigger strong emotions in patients suffering from childhood trauma. Should patients be uncomfortable with the suggested protocol at any point, they are given the option to decline treatment. The application will also take note of the protocol and omit it in the remainder of the session.
\end{enumerate}

It should also be stressed that the SAT chatbot, while a mental health application, is not equipped, without a concurrent intervention by a human psychotherapist, to treat patients suffering from serious mental health conditions such as severe depression. Participants are only permitted to take part in the human evaluation trials once they have been thoroughly informed of the associated risks, and clear consent has been received.

\subsection{Data Protection}
While patients do not need to provide personal information to interact with the chatbot, the \textit{contents} that patients discuss with the chatbot are themselves considered personal data under the UK's Data Protection Act (DPA) \cite{DPA2020DataProtection} and General Data Protection Regulations (GDPR) \cite{GDPR2020General}. To ensure adherence to data protection laws, the SAT chatbot does not store user interactions beyond the treatment sessions. We also do not store metadata from user's devices (e.g. geolocation, IP/MAC addresses, IMEI codes etc.).

Data compliance was also ensured during the human trial. The human trial conducted in this paper has been approved by the Imperial College Research Ethics Committee. Prior to the trial, participants were informed on how their personal information would be handled, and were required to provide consent before participating. Moreover, participant responses are anonymised, and all responses collected are used strictly for the purposes of the current study.

%% file: sections/III-dataset.tex
\subsection{Data Analysis}
\label{dataset:data_analysis}
The \texttt{EmpatheticPersonas} dataset \cite{Alazraki2021Satbot} (EP) is a crowd-sourced dataset intended for the development of the SAT chatbot. This dataset is comprised of two main components. Firstly, it contains 1,181 written expressions of \textbf{emotion}, aimed at training an emotion classifier. The examples in the dataset are approximately evenly distributed across four emotion classes: there are 284 examples relating to Fear/Anxiety, 297 relating to Anger, 300 relating to Sadness and 300 relating to Joy/Contentment. Secondly, the dataset contains 2,144 \textbf{empathetic rewritings} of 45 base utterances. 1,100 of these have also been annotated for empathy using a discrete scale from 0 to 2 (where 0 represents non-empathetic utterances, 1 represents slightly empathetic ones and 2 corresponds to highly empathetic utterances). The annotated rewritings are aimed at training an empathy classifier.

We produced an additional native Mandarin dataset consisting of 120 emotional utterances balanced across four emotion classes for testing purposes.

\subsection{Dataset Translation}
\label{dataset:translation}
\begin{figure}[tb]
    \centering
    \includegraphics[width=\linewidth]{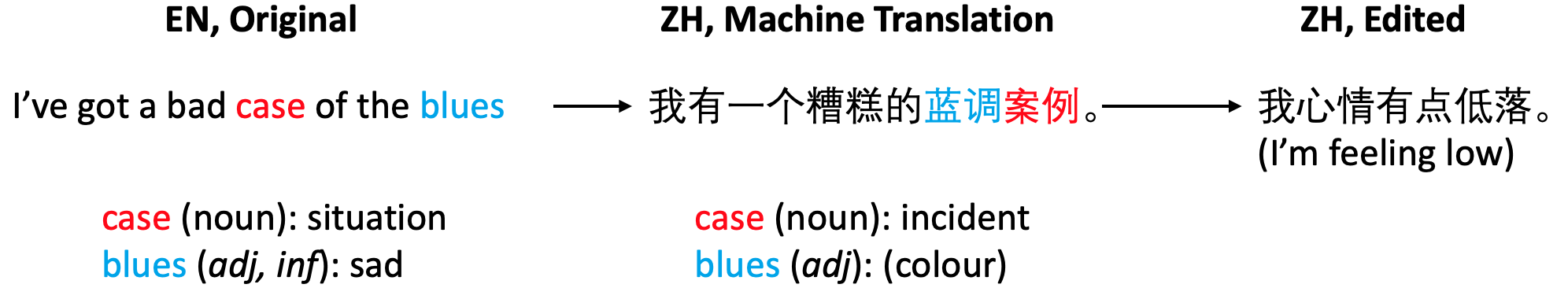}
    \caption{Example of major translation errors targeted in post-editing. Most translation errors stemmed from literal translations of EN colloquialisms into ZH.}
    \label{fig:dataset:translation1}
\end{figure}

\begin{figure}[tb]
    \centering
    \includegraphics[width=0.9\linewidth]{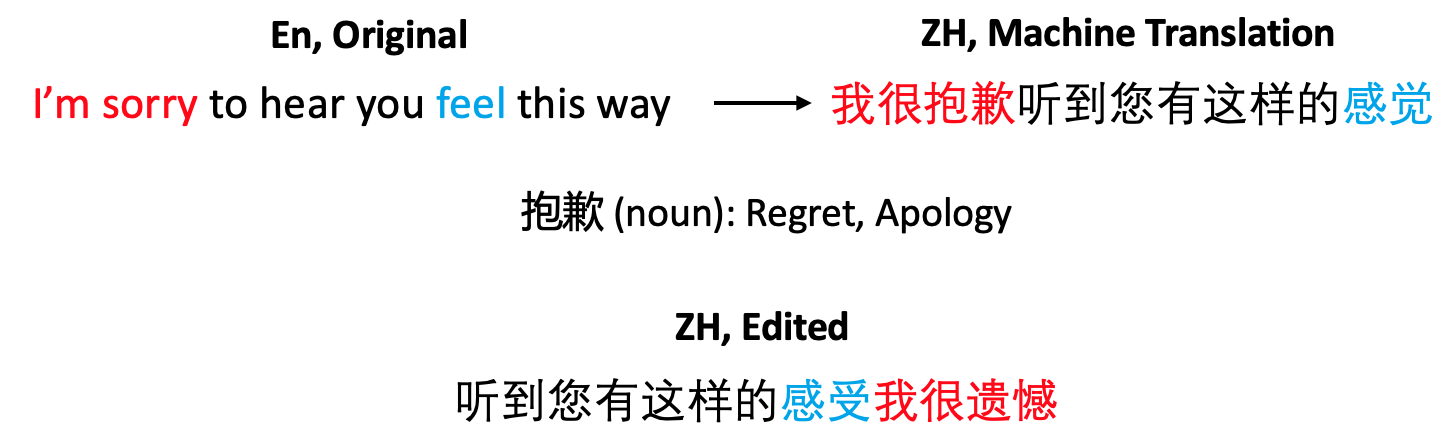}
    \caption{Example of minor translation errors edited to increase localisation accuracy. Note that the original translation maintained the coherence of the source sentence, but used words that were not the most appropriate to the context.}
    \label{fig:dataset:translation2}
\end{figure}

As crowd-sourcing data is a time consuming and costly process, we leveraged publicly available machine translation tools to aid the translation process of the existing EP dataset into Mandarin. We formulated our translation pipeline as follows:

\begin{enumerate}
    \item We used a publicly available machine translation tool (Google Translate) to obtain a base English(EN)-Mandarin(ZH) translation of the EP dataset.
    \item We performed post-editing (\texttt{v1}) on the translated dataset to remedy major translation errors affecting sentence semantics (Fig. \ref{fig:dataset:translation1}).
    \item An additional post-editing step (\texttt{v2}) was introduced after early trials identified a need to inject language-specific terms and colloquialisms to further improve the localisation quality of the translations (Fig. \ref{fig:dataset:translation2}).
\end{enumerate}


It is worth noting that the introduction of post-editing steps allows screening of candidate responses for potentially harmful or dangerous utterances in addition to remedying errors.

Reference-based sentence evaluation metrics such as BLEU\cite{Papineni2002Bleu} and ROUGE\cite{Lin2004Rouge} evaluate the quality of a translation against reference target sentences. As human-translated target sentences were not available, we instead evaluated the efficacy of our translation using the reference-free \cite{Ethayarajh2020BleuNeighbours} sentence fluency metrics SLOR \cite{Kann2018Fluency} and PRSIM-SRC\footnote{We show NLL in our work as opposed to the Log-likelihood shown in the original work \cite{Thompson2020Automatic}.}\cite{Thompson2020Automatic} along with sentence perplexity (PPL), with results shown in Table \ref{tab:dataset:translation1}.

\begin{table}[bt]
\def\arraystretch{1.25}
\begin{center}
\caption{Average sentence SLOR scores (higher is better), PRISM-SRC scores (lower is better) and Perplexity (PPL) scores (lower is better) for different revisions of the dataset.}
\begin{tabular}{| l | c | c | c |}
    \hline
    \textbf{Revision} & \textbf{SLOR} & \textbf{PRISM-SRC} & \textbf{PPL} \\
    \hline
    \texttt{Base} & 3.84 & 39.05 & 19.77 \\
    \texttt{\texttt{v1}} & 3.87 & \textbf{34.71} & \textbf{18.08} \\
    \texttt{\texttt{v2}} & \textbf{3.92} & 35.04 & 18.83 \\
    \hline
\end{tabular}
\label{tab:dataset:translation1}
\end{center}
\end{table}

We observe from Table \ref{tab:dataset:translation1} that both post-edit revisions yield better fluency scores across all three metrics, suggesting that the inclusion of post-edits did improve the fluency of the utterances over base machine translated text, and improved translation quality with respect to the source sentence. We note that \texttt{v2} yields a slightly higher improvement over the \texttt{base} version in SLOR (3.92 vs 3.84) than \texttt{v1} (3.87 vs 3.84), whilst \texttt{v1} scores higher on PRISM-SRC (34.71 vs 35.04) and Perplexity (18.08 vs 18.83). We hypothesise that this may be due to the fact that the edits made in \texttt{v2} are of a finer-grain nature, using `rarer' tokens that are more commonly associated with colloquialisms.

We also compare the quality of our utterances to the English version through human evaluation in Section \ref{trial:evaluation}, where we show that our post-edited utterances attain a comparable level of user experience to the English version.

%% file: sections/IV-implementation.tex
\begin{figure}[tb]
\centerline{\includegraphics[width=0.4\linewidth]{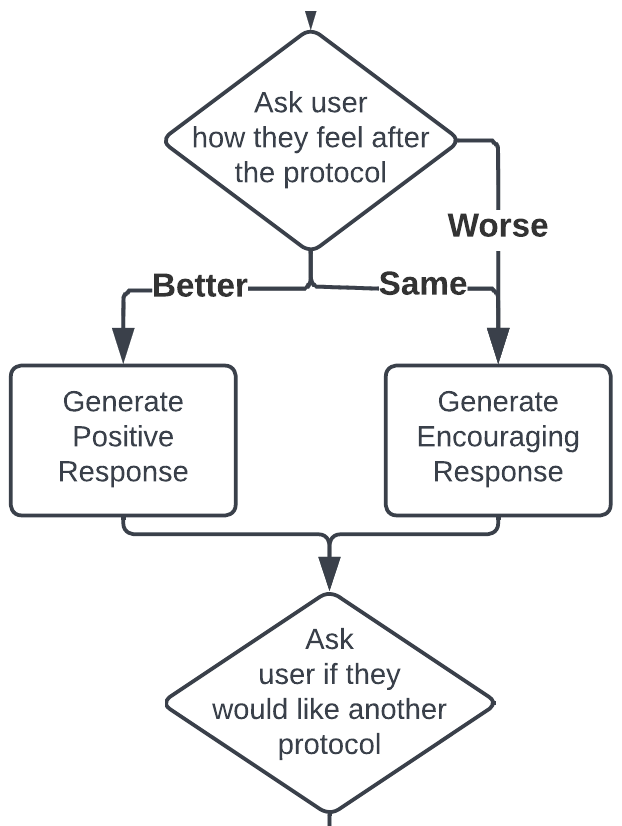}}
\caption{Updated conversation flow component for producing empathetic responses based on the user's change in emotion (after practising a protocol) and their original emotional state.}
\label{fig:conv-flow}
\end{figure}

Our chatbot uses a rule-based conversation flow as in~\cite{Alazraki2021Satbot}, where the chatbot first works to recognise the user's emotional state and from this it guides the next stages of the conversation to establish a context for protocol recommendation, with previously ineffective protocols removed from the set of potential recommendations. We introduce an addition to the conversation flow as shown in Fig.~\ref{fig:conv-flow}, to account for the user's change in emotional state after carrying out a protocol and produce appropriate responses. The chatbot contains two core components: \textit{emotion recognition} and \textit{empathetic rewritings}.

\subsection{Emotion Recognition}
\label{section:emotion_recognition}
As the conversation flow is dictated by the user's emotional state, the chatbot needs to correctly identify the user's emotions. We developed an emotion classifier capable of identifying four emotions: sadness, anger, fear/anxiety and joy/contentment. We double finetuned the pretrained language model (PLM) XLM-R\footnote{Available at \url{https://huggingface.co/xlm-roberta-base}}\cite{Alexis2019XLMR} using an emotion dataset in native Mandarin (NLPCC)\cite{Zhou2018Emotional}, followed by the EP emotion data. Our model's results are shown in Table \ref{table: emo-results}, where at least 90\% accuracy and F1-scores are attained across all test sets. The first finetuning was introduced to enhance performance in native Mandarin. However, our findings show that the model performs sufficiently well even without finetuning on the NLPCC dataset (see `single' results in Table \ref{table: emo-results}). This is beneficial for low resource languages where there may be a lack of native in-domain data. 

\begin{table}[tb]
\def\arraystretch{1.25} 
\setlength{\tabcolsep}{12pt} 
\caption{Emotion classifier results against different \texttt{EmpatheticPersonas} test sets with single and double finetuning.}
\begin{center}
\begin{tabular}{|c|c|cc|}
\hline
&     & \textbf{Accuracy} & \textbf{F1-Score} 
\\ \hhline{~~--} 
\multirow{-2}{*}{\textbf{Test Set}} & \multirow{-2}{*}{\textbf{Finetuning}} & \textbf{\%}            & \textbf{\%}       \\ \hline
ZH              & double   & 93.86    & 93.86    \\
(translated)   & single   & 92.98    & 93.02    \\ \hline
ZH              & double   & 90.00    & 90.09    \\
(native)        & single   & 84.17    & 83.86    \\ \hline
                & double   & 91.23    & 91.40    \\
\multirow{-2}{*}{EN}       & single   & 89.47    & 89.53    \\ \hline
\end{tabular}
\label{table: emo-results}
\end{center}
\end{table}

\subsection{Knowledge Distillation}
\label{implementation:knowledge_distillation}
While large PLMs have allowed for state-of-the-art performance in various NLP tasks, their size makes them computationally expensive and memory intensive to operate. Hence, adopting such models in real time applications becomes highly impractical due to cost and latency issues \cite{Sanh2019Distilbert}.

To optimise the emotion classifier for runtime efficiency, we performed Knowledge Distillation \cite{Hinton2015Distilling} as a compression technique to reduce the size of the model while maintaining its performance. Using the double finetuned model (Section \ref{section:emotion_recognition}) as the teacher model, we performed Knowledge Distillation on a L6xH384 mMiniLMv2 student model\footnote{Available at \url{https://github.com/microsoft/unilm/tree/master/minilm}}\cite{Wenhui2020MiniLMv2}, a task-agnostic model distilled from an XLM-R-large model. We performed double finetuning, with distillation occurring at each stage, inspired by the multi-stage distillation framework shown in \cite{Jiao2020Tinybert}.

Distillation was performed using the Triple Loss method \cite{Sanh2019Distilbert}, which incorporates distillation loss (\(L_{dist}\)) and cosine embedding loss (\(L_{cos}\)), in addition to the classic supervised training loss (\(L_{ce}\)), during training.
\hfill \break

\begin{enumerate}
    \item \textbf{Classic Supervised Training Loss, \(L_{ce}\)} \\
        This is the cross-entropy loss between the student model's predicted distribution (\(c_i\)) and the target training labels (\(q_i\)) which is in the form of a one-hot vector.
        \begin{equation}
            L_{ce} = \sum_i q_i * log(c_i)
        \end{equation}
     
    \item \textbf{Distillation Loss, \(L_{dist}\)} \\
        This is the cross-entropy loss between the student model's  \textit{softened} predicted distribution (\(s_i\)) and the teacher's  \textit{softened} predicted distribution (\(t_i\)) \cite{Hinton2015KD}. 
        \begin{equation}
            L_{dist} = \sum_i t_i * log(s_i)
        \end{equation}
        
        These softened predictions are also known as the softmax-temperature probability distribution, given by:
        \begin{equation}
            p_i = \frac{exp(z_i/T)}{\sum_j exp(z_j/T)}
        \end{equation}
        
        where $T$ denotes temperature and $z_i$ denotes the probability of class $i$.
        \hfill \break
      
    \item \textbf{Cosine Embedding Loss, \(L_{cos}\)} \\
        While most Knowledge Distillation methods use only losses 1 and 2, the cosine embedding loss is specific to Triple Loss. It aims to align the student's and teacher's hidden vector representations and is noted to improve performance \cite{Sanh2019Distilbert} . The loss is as follows:
        \begin{equation}
            L_{cos} = 1 - cos(T(x),S(x))
        \end{equation}
\end{enumerate}

Thus, the final training loss is taken as the average of the three losses:
\begin{equation}
    L_{total} = \frac{L_{ce} + L_{dist} + L_{cos}}{3}
\end{equation} 

\begin{figure}[tb]
\centerline{\includegraphics[width=1.0\linewidth]{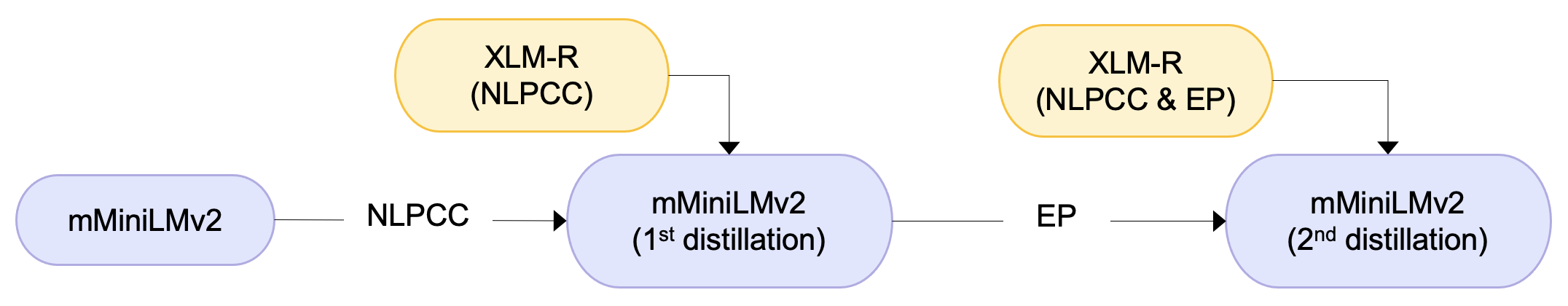}}
\caption{Model training pipeline for the emotion classifier. It involves a two stage finetuning, with distillation occurring at each stage. Teacher models (XLM-R) are represented in yellow and student models (mMiniLMv2) in blue.}
\label{fig:distillation}
\end{figure}

The emotion classifier training pipeline is illustrated in Fig.~\ref{fig:distillation}. Following hyperparameter tuning, we obtained a performance (accuracy and F1) of $\sim$81\% and $\sim$85\% on the native Mandarin and English test sets respectively (Table \ref{table: final-distilled}). Considering that the mMiniLMv2 has only 40\% of the XLM-R-base teacher model's capacity (see Table \ref{table: model-size}), the model performs extremely well, retaining a significant proportion of its teacher model's performance (approx. 90\% in the worst case and up to 97\% in the best case.). We also note a significant reduction in average inference time using the distilled model (Table \ref{table:inference-time}) compared to the base XLM-R model. This had a noticeable impact on trial participant feedback (see Section \ref{trial:evaluation}). 

\begin{table}[bt]
\def\arraystretch{1.25} 
\caption{Accuracy and F1-scores of the distilled student model (mMiniLMv2) on the 3 EP test sets. We also include the performance of the teacher model (XLM-R) and the English emotion classifier from \cite{Alazraki2021Satbot} for comparison.}
\begin{center}
\begin{tabular}{|c|cc|cc|cc|}
\hline
& \multicolumn{2}{c|}{\textbf{ZH Translate}} & \multicolumn{2}{c|}{\textbf{ZH Native}} & \multicolumn{2}{c|}{\textbf{EN}} \\  \hline
& \textbf{Acc.}  & \textbf{F1}  & \textbf{Acc.}  & \textbf{F1}  & \textbf{Acc.}  & \textbf{F1} \\ \hhline{~------} 
\multirow{-2}{*}{\textbf{Model}} & \textbf{\%}  & \textbf{\%}  & \textbf{\%}  & \textbf{\%}  & \textbf{\%}  & \textbf{\%} \\ \hline

RoBERTa-base \cite{Alazraki2021Satbot} & -& - & - & - & 94.96 & 95.10  \\ \hline\hline
XLM-R-base  & 93.86  & 93.86  & 90.00  & 90.09  & 91.23   & 91.40    \\
mMiniLMv2  & \underline{91.22}  & \underline{91.31}  & \underline{80.83}  & \underline{80.85}  & \underline{85.09}  & \underline{85.39}                  \\ \hline
\end{tabular}
\label{table: final-distilled}
\end{center}
\end{table}

Additionally, we compared the distilled model's performance against the previous SAT chatbot emotion classifier (see Table \ref{table: final-distilled}). When comparing against the English RoBERTa-base classifier deployed in \cite{Alazraki2021Satbot}, the distilled model has 90\% of its performance at 40\% of its capacity (Table \ref{table: model-size}). On the other hand, the non-clinical trial results point to a significant improvement in participant sentiment toward our emotion classifier compared to the one from \cite{Alazraki2021Satbot}.

Overall, considering the computational advantages and minimal performance trade-off, the above results illustrate the potential of performing Knowledge Distillation, whereby classification performance can be largely recreated with a significantly smaller and more efficient model.

\begin{table}[tb]
\centering
\def\arraystretch{1.25} 
\caption{Model size comparisons.}
\begin{tabular}{|l|c|c|c|}
\hline
\textbf{Model} & \textbf{Layers} & \textbf{Hidden Dim.} & \textbf{No. Params.} \\
\hline
XLM-R \cite{Alexis2019XLMR} & 12 & 768 & 270M \\
mMiniLMv2 \cite{Wenhui2020MiniLMv2} & 6 & 384 & 107M \\ 
\hline
\end{tabular}
\label{table: model-size}
\end{table}

\begin{table}[tb]
\centering
\def\arraystretch{1.25} 
\caption{Average inference time.}
\begin{tabular}{|l|c|}
\hline
\textbf{Model} & \textbf{Inference Time (s)} \\ \hline
XLM-R-base & 0.1877 \\
mMiniLMv2 &  0.0308 \\ 
\hline
\end{tabular}
\label{table:inference-time}
\end{table}

\subsection{Empathetic Rewriting}
\label{implementation:empathetic_rewriting}

In order to increase the level of empathy expressed in the chatbot's responses, we augmented the existing translated responses by having lower-empathy utterances rewritten to be more empathetic. As the chatbot has a rule-based conversational flow, rewriting has the additional benefit of boosting diversity in its conversation, thus potentially leading to greater user engagement.

We adopted the generative language model Chinese GPT-2\footnote{Available at \url{https://huggingface.co/uer/gpt2-chinese-cluecorpussmall}} to generate the empathetic rewritings in Mandarin. This model was trained using reinforcement learning (RL) with proximal policy optimisation \cite{Schulman2017Proximal}, based on \cite{Ziegler2019FineTune}. Prior to training, we performed a supervised warm-start since literature has shown that it leads to more effective learning \cite{Sharma2021Empathic, Ziegler2019FineTune}. 

\begin{figure*}[tb]
    \centering
    \includegraphics[width=0.9\textwidth]{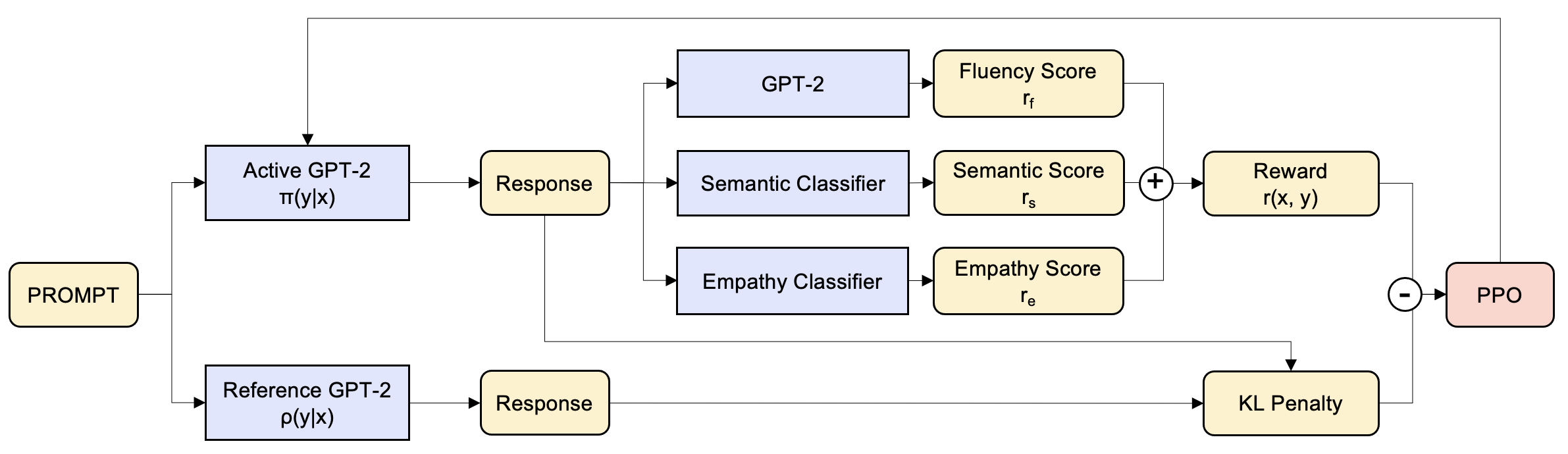}
    \caption{Reinforcement learning setup for empathetic rewriting.}
    \label{fig:rl_setup}
\end{figure*}


The training setup is illustrated in Fig. \ref{fig:rl_setup}. To facilitate training, we devised a reward model to reward utterances that are first and foremost empathetic, but also fluent and semantically relevant. 

The \textbf{empathy reward} \( r_e \)
is the key component of the empathetic rewriting task. This component aims to reward rewritings that convey a high degree of empathy, and penalise rewritings conveying low empathy. To quantify the degree of empathy conveyed by an utterance, we developed an empathy classifier using an XLM-R model trained and evaluated on the empathy-annotated EP data, obtaining an overall accuracy and F1-score of 90\%. The logit of the highly empathetic class computed by the classifier is then taken as the empathy reward.

The \textbf{semantic reward} \( r_s \)
aims to reward rewritings that deliver the same semantic meaning as the base utterance. Without this component, utterances that are highly empathetic but do not carry the correct semantic information may be generated as the model seeks to exploit the empathy reward. To measure the semantic similarity of a rewriting to its base utterance, we trained and evaluated an XLM-R model on the empathetic rewritings in the EP dataset, obtaining an overall accuracy and F1-score of 96\%. The semantic reward is the logit of the semantic class corresponding to the base utterance.

The \textbf{fluency reward} \( r_f \)
was adapted from the fluency function in \cite{Alazraki2021Satbot} and included to prevent rewritings that are highly empathetic but are incoherent/grammatically incorrect. This is computed as:
\begin{equation}
    r_f(er) = \frac{1}{PPL(er)} - RP(er)
\end{equation}
where $er$ denotes the empathetic rewriting, $\frac{1}{PPL(er)}$ is the inverse of the perplexity (computed by a GPT-2 model) and $RP(er)$ denotes a cumulative penalty for every repeated word within that rewriting (excluding stop words). Attempting to remove the repetition penalty term resulted in the model seeking to exploit the semantic and empathy reward by repeating keywords/empathetic terms.

The final reward was implemented as a multi-objective function comprised of the weighted sum of the empathy, fluency and semantic rewards, written as:
\begin{equation}
    r = w_{e}r_{e} + w_{f}r_{f} + w_{s}r_{s}
\end{equation}

Similarly to related works \cite{Nieminen2022Coproducing}, we pre-generated and manually inspected empathetic responses for any toxic speech or distressing content (e.g. relating to violence or self harm) before approving them to be used by the chatbot. It is worth noting, however, that no problematic content was found in the utterances generated by the final trained model. In the future, a hate-speech detector could be devised to automate this inspection process.

\subsection{Supervised Empathetic Rewriting}
\label{implementation:supervised}
While the RL-based methodology yielded overall quality responses, we should note that this method can be extremely sensitive. As PPO is a stochastic policy method, its actions are drawn from a probability distribution. This means that actions vary each time, resulting in starkly different outcomes between different runs of training. Moreover, performance varies significantly based on the weights attached to the reward components (i.e., $w_e$, $w_s$ and $w_f$), which makes hyperparameter tuning difficult.

In response to this, we also introduce a simpler, supervised learning (SL) approach to empathetic rewriting. We fine-tuned a GPT-2 model by prompting it with the user's emotional state $\mathcal{S}_e$ and a basic, low-empathy utterance $\mathcal{S}_L$, and used a high-empathy utterance $\mathcal{S}_H$ as the learning target. We employed the empathy classifier (EC) used in the RL method to form an additional binary classification learning objective over all the utterances $X_g$ generated at each training step:

\begin{equation}
    L_{EC} = CrossEntropyLoss(EC(X_g), 1)
\end{equation}

Where $1$ is the label for highly empathetic sentences. We then updated the model using the combined loss

\begin{equation}
    L_{Total} = L_{LM} + L_{EC}
\end{equation}
where $L_{LM}$ is the language modelling loss produced by the GPT-2 model.

We then pre-generated and manually inspected responses in the same manner as the RL approach. We compare the responses generated by the two approaches through non-clinical human trials against the English SAT chatbot \cite{Alazraki2021Satbot} in Section \ref{trial:evaluation}.

%% file: sections/V-trial.tex
\subsection{Study Design}
\label{trial:study_design}
Formal evaluation of the SAT chatbot was carried out through non-clinical human trials. Separate trials were conducted on chatbots using responses generated via the reinforcement learning approach and the supervised learning method. For the purpose of the trial, participants were required to be fluent in both Simplified Mandarin and English in order to fully experience the bilingual chatbot. We note that users fluent in either language are nonetheless able to utilise the chatbot to practise SAT. Given the limited participant pool, knowledge of SAT protocols or psychotherapy was beneficial but not required. Participants were nonetheless provided with information detailing the SAT protocols prior to the trials. In total, 42 participants (20 female, 22 male) aged 25 to 60 consented to and participated across three trials.

Throughout each trial, participants were instructed to interact with the application once per day over a period of five days. There had to be a minimum of three interactions in Mandarin and one in English. Participants were also asked to note down any unnatural sounding utterances generated by the chatbot when using the Mandarin setting.

At the end of each trial, an anonymous feedback questionnaire was issued to each participant, aimed at evaluating their experience using the chatbot. The questionnaire sought to collect user feedback on: (i) the chatbot's emotion recognition capabilities, (ii) the quality of and the empathy conveyed by the chatbot's responses, (iii) the overall experience of using the chatbot and (iv) the perceived usefulness of the chatbot. Participants were asked to evaluate each aspect by providing their level of agreement with a particular statement on a Likert scale.

\subsection{User Interface}
\label{trial:user_interface}

\begin{figure}[tb]
\centerline{\includegraphics[width=\linewidth]{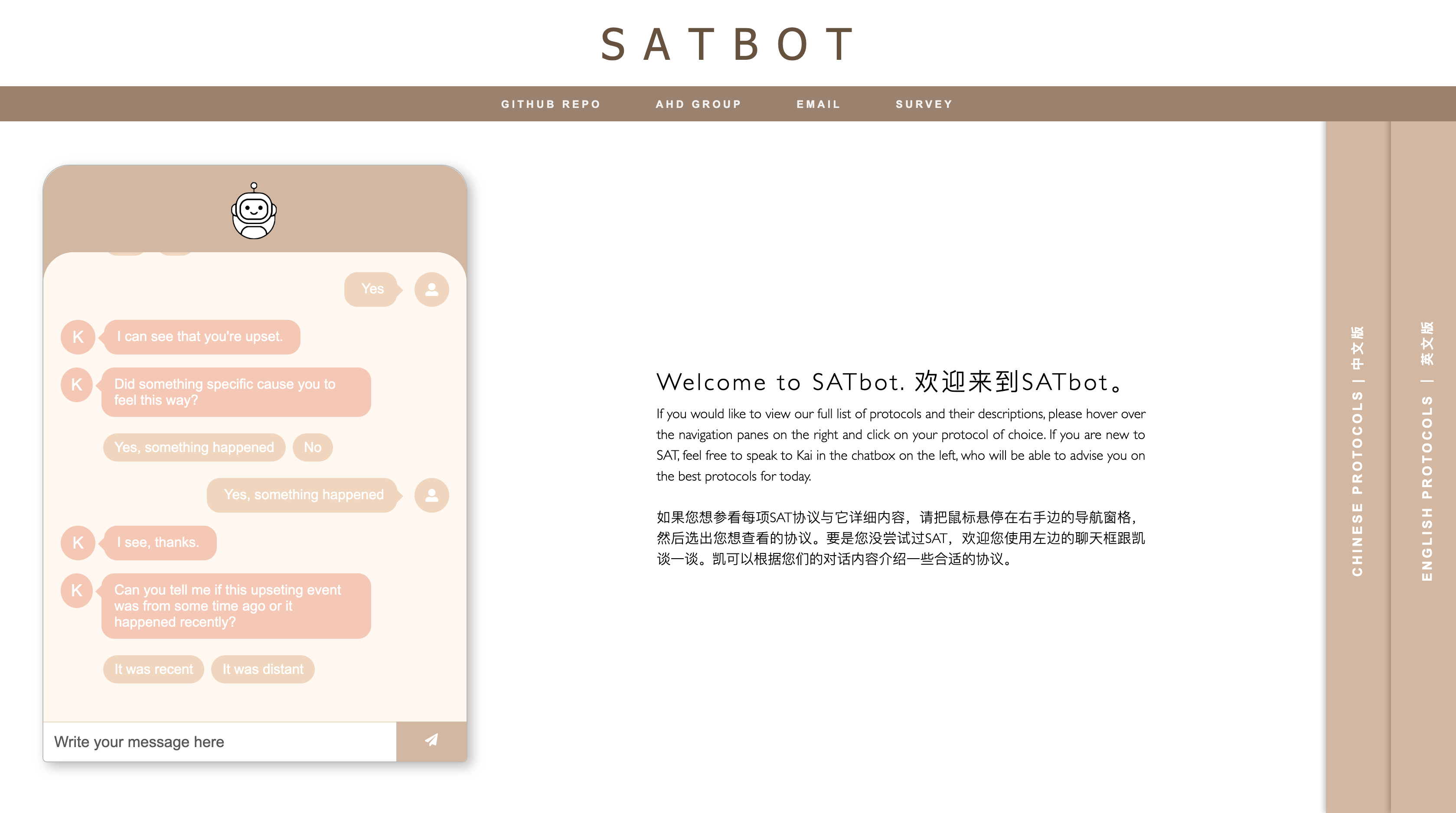}}
\caption{Trial platform web interface.}
\label{fig:satbot-interface}
\end{figure}

Fig. \ref{fig:satbot-interface} shows the user interface of the web platform that was deployed for the non-clinical trial. Protocols were available to view on the platform upon selection in both English and Mandarin.

\subsection{Evaluation}
\label{trial:evaluation}
\begin{figure}[t]
\centerline{\includegraphics[width=0.78\linewidth]{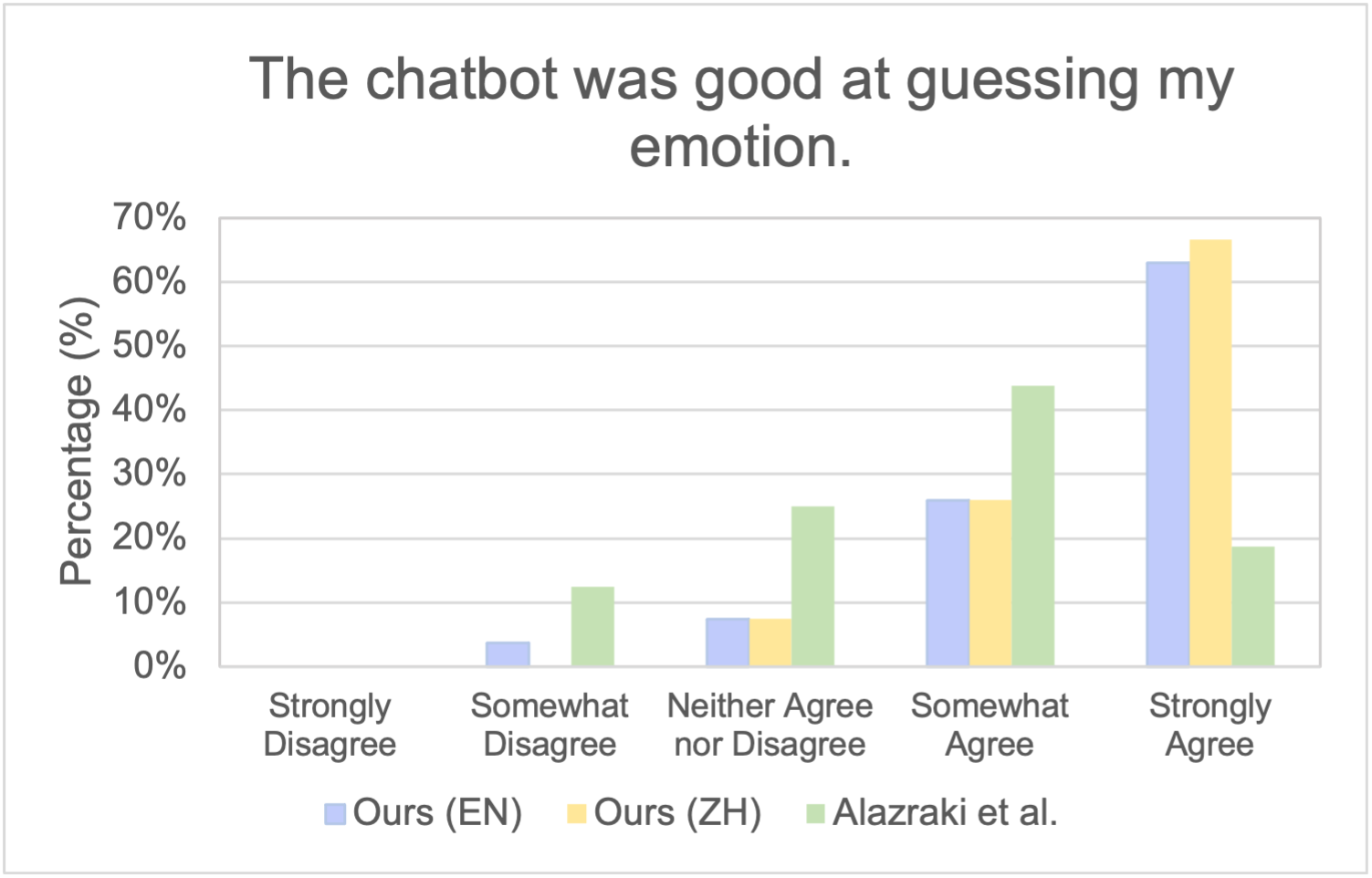}}
\caption{Participant evaluation of emotion classification performance.}
\label{fig:trial-emotion}
\end{figure}

Participants were first asked to evaluate the emotion classifier's capabilities. When assessing whether the chatbot was good at guessing emotions, 89\% and 93\% of participants agreed with this statement for the emotion classification in English and Mandarin respectively compared to previous works \cite{Alazraki2021Satbot}, where only 63\% of participants agreed (see Fig. \ref{fig:trial-emotion}). Taking into account that the distilled model is half the size of the one used in \cite{Alazraki2021Satbot}, this highlights the success of Knowledge Distillation at achieving performant yet compact models.

\begin{figure}[tb]
\centerline{\includegraphics[width=0.78\linewidth]{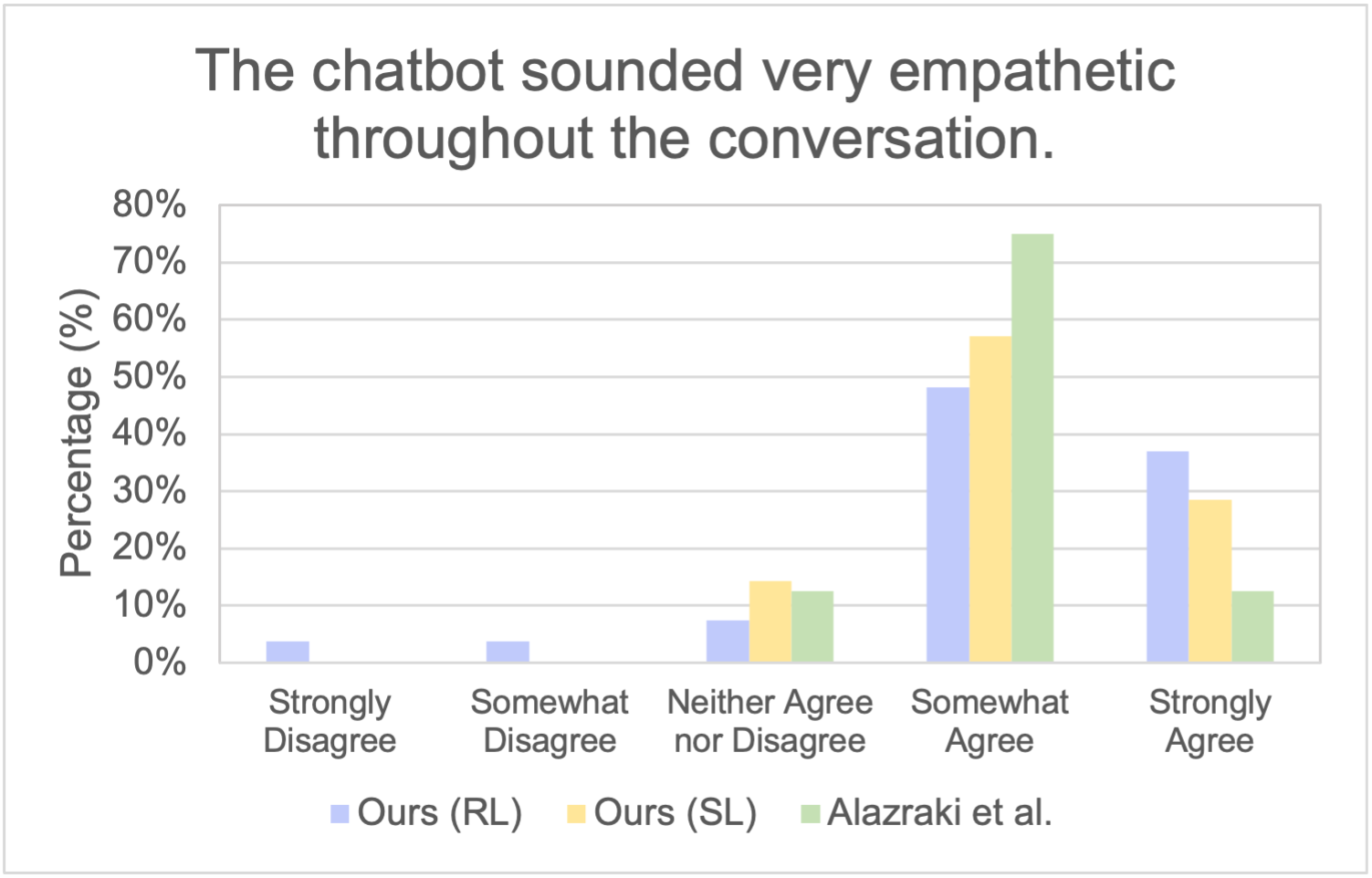}}
\caption{Participant evaluation of the chatbot's display of empathy during conversation.}
\label{fig:trial-empathy}
\end{figure}

Participants were also asked to evaluate the quality of the chatbot's utterances. When asked to gauge whether the chatbot came across as highly empathetic throughout the conversation, 85\% of the participants that had interacted with the RL chatbot agreed, while this proportion was 86\% for the participants that had interacted with the SL chatbot (see Fig. \ref{fig:trial-empathy}). This result is consistent with the previous English-only implementation, where 88\% of participants had agreed with this statement.

\begin{figure}[tb]
\centerline{\includegraphics[width=0.78\linewidth]{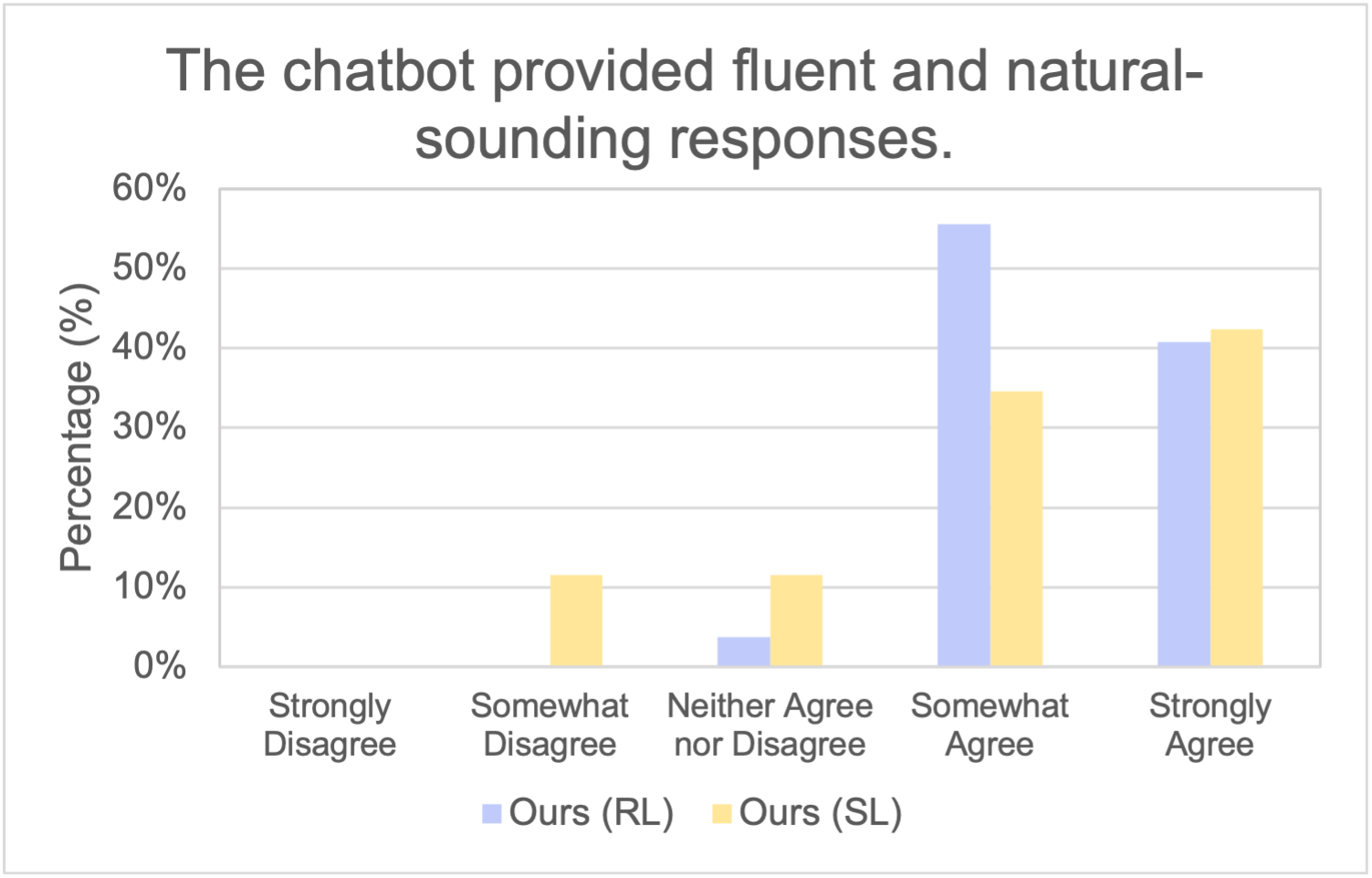}}
\caption{Participant evaluation of the chatbot's fluency and naturalness in conversation.}
\label{fig:trial-fluent}
\end{figure}

When asked if they found that the chatbot provided fluent and natural-sounding responses, 96\% of participants that had used the RL chatbot agreed, while this proportion was 77\% for the participants that had engaged with the SL chatbot (see Fig. \ref{fig:trial-fluent}). 

\begin{figure}[tb]
\centerline{\includegraphics[width=0.78\linewidth]{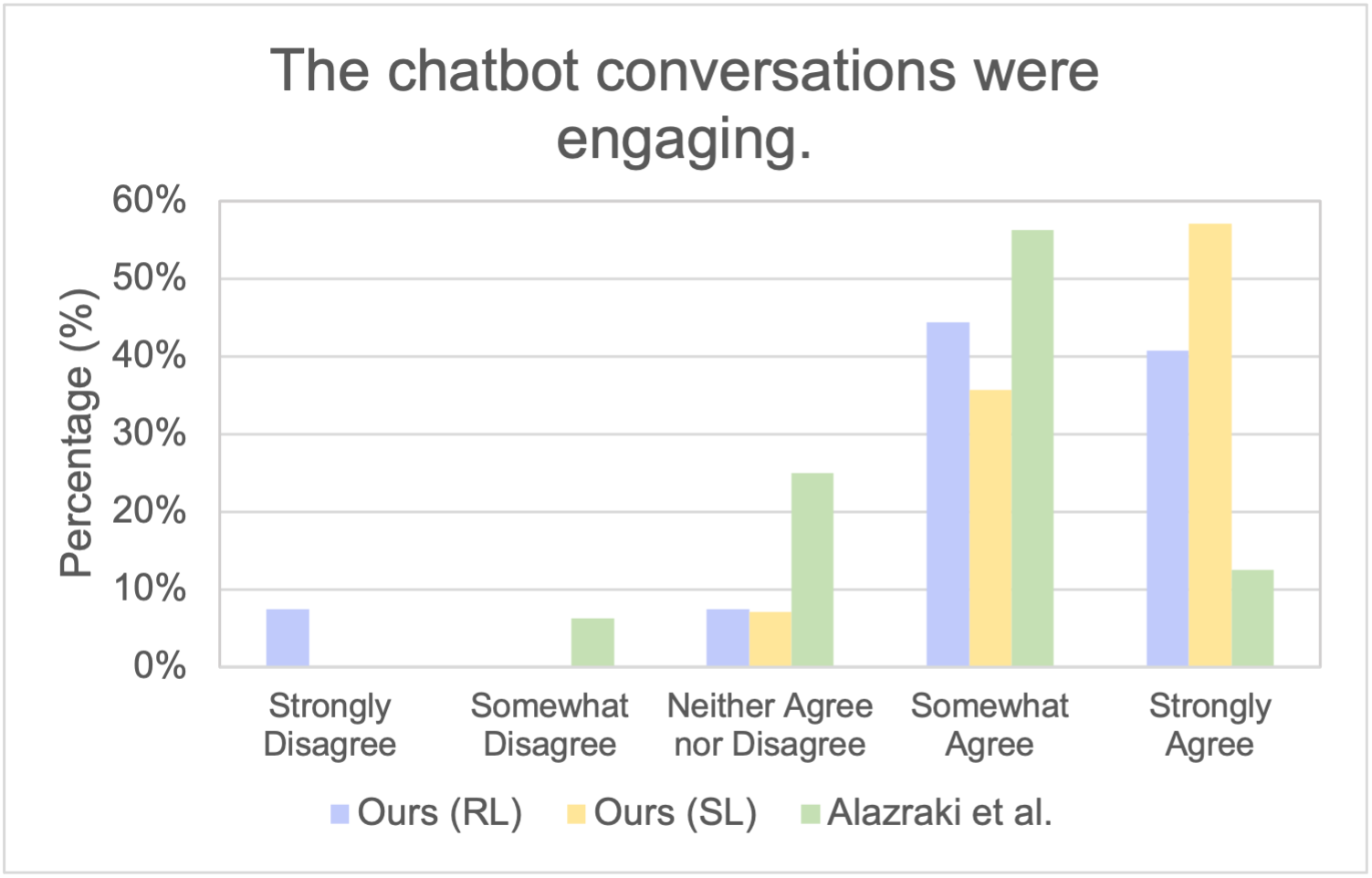}}
\caption{Participant evaluation of the chatbot's engagement.}
\label{fig:trial-engagement}
\end{figure}

With regards to the level of engagement when conversing with the chatbot, 85\% of participants agreed that they were engaged when the chatbot used RL-trained utterances, while 93\% agreed when the chatbot used SL-trained utterances (see Fig. \ref{fig:trial-engagement}). The perceived user engagement of our platform is thus significantly higher than the previous English-only implementation, where only 69\% of trial participants had agreed with the above statement.

\begin{figure}[tb]
\centerline{\includegraphics[width=0.78\linewidth]{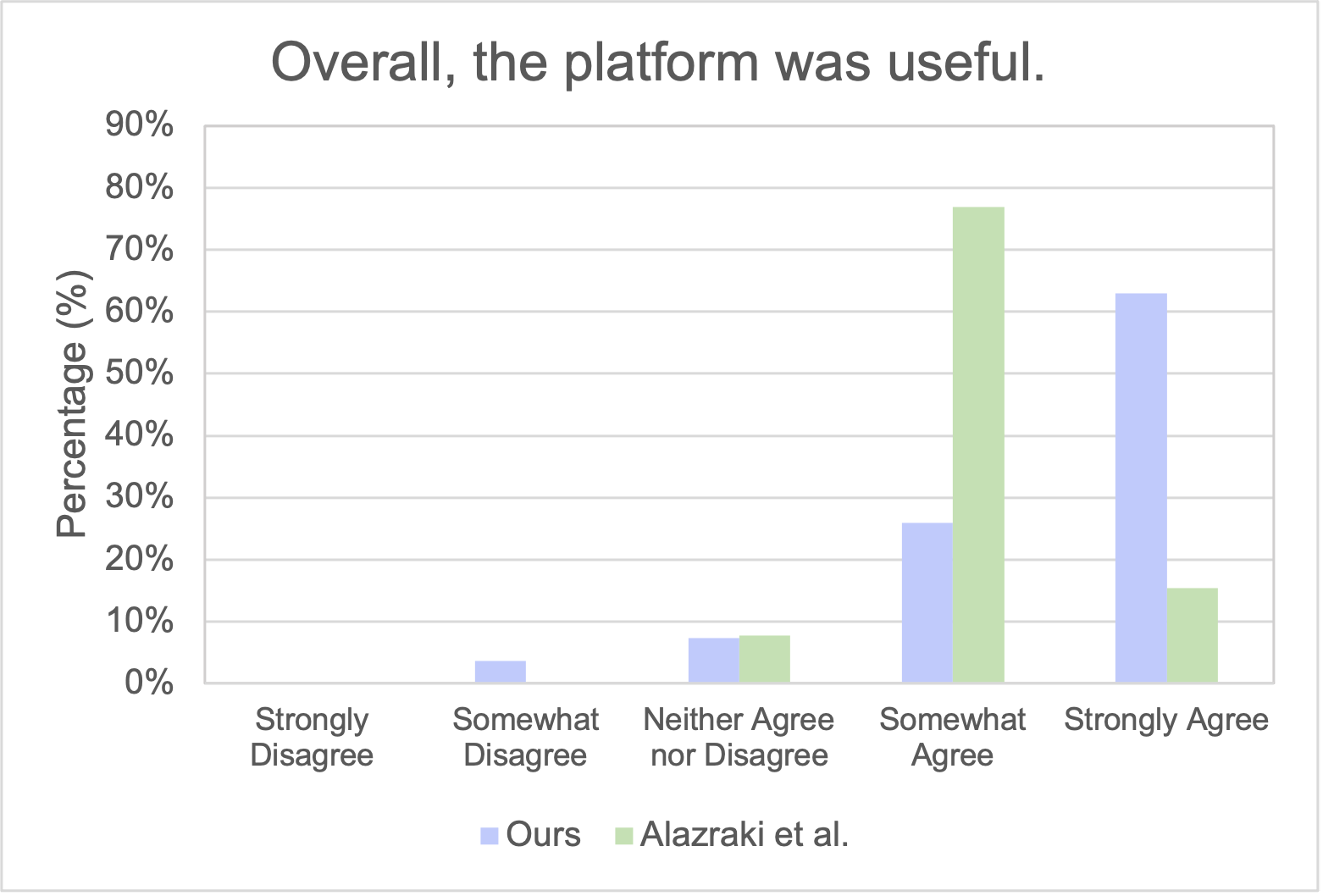}}
\caption{Participant evaluation of the chatbot's usefulness.}
\label{fig:trial-useful}
\end{figure}

Finally, participants were asked to evaluate the chatbot's usefulness. 89\% of participants agreed that the platform was useful. This is roughly consistent with the proportion of users who had agreed that the English-only platform was useful, which was 92\% (see Fig. \ref{fig:trial-useful}).

%% file: sections/VI-discussion.tex
\subsection{Results}
\label{discussion:results}

Our framework builds upon previous work to provide a foundation for future multilingual development and for the deployment of computational methods for mental health support to more languages, with a more scalable pipeline and less reliance on in-domain, in-language data. The human trials of our platform have shown promising results with respect to the perceived empathy, usefulness, user engagement, quality of responses and ability to correctly identify users' emotions.

\subsection{Study Limitations}
\label{discussion:limitations}
The key limitation faced in this study is that the majority of participants did not possess knowledge of SAT protocols prior to the trial. This was a necessary trade-off as participants were required to be English-Mandarin bilingual speakers, which significantly limited the size of the participant pool. Efforts were made to inform participants on how to carry out SAT protocols prior to the start of the human trial through information documents, and we received positive feedback from some participants regarding SAT. Nonetheless, the results from the trial evaluation, such as those pertaining to empathy and usefulness of the chatbot, may not be entirely reflective of the chatbot's performance from a psychology-specific point of view. In the future, trial participants should undergo training in SAT where they can receive information on the treatment and its protocols, for a better understanding before providing feedback. If possible, having Mandarin-speaking clinicians participating in the trials would also be extremely valuable.

Another notable limitation was the study size. Whilst our trial recruited more participants than the previous study \cite{Alazraki2021Satbot}, the participant sample was still relatively small. Moreover, the study sizes across the three trials conducted were inconsistent (13, 14 and 27), with imbalanced demographics across the sexes. This is once again due to the stringent requirements of the trial screening which limited the participant pool. In future trials, recruitment should continue to increase the trial sample size and focus on balancing demographics. \\


\subsection{Future Work}
\label{discussion:future_work}

We note that the human trial was conducted with the purpose of quantifying the efficacy of using the multilingual chatbot, and was not aimed at determining the therapeutic effects of SAT on a Chinese-speaking population. An 8-week psychological intervention can be conducted in the future, where participants are exposed, step by step, to the SAT protocols through weekly sessions, and in which a Mandarin-capable SAT chatbot can be used in guiding users through carrying out SAT protocols on a daily basis. Future work could also investigate the application of this translation-based method to the delivery of other rule-based psychotherapy methods, such as CBT, on a Chinese-speaking population.

Code-switching, sometimes referred to as code-mixing, is a phenomenon prevalent in multilingual communities, whereby individuals alternate between two or more languages within the same conversation \cite{Sebastin2021codeswitch}. Since a potential input to the chatbot could contain code-switched text, it would be interesting to see how model performance can be optimised for such inputs.

Moreover, it would be worth investigating the performance of the same model used in this paper on different languages, especially those with low resource availability. An extension could also be designed for expanding the range of emotions recognised by the classifier, and to assess the efficacy of formulating the emotion classification task as a multi-label problem, as human emotions can be complex and are typically not mutually exclusive.

A common reflection amongst trial participants focuses on the inherent rigidity of rule-based conversation. Therefore, future work could investigate the incorporation of open-dialogue to facilitate more natural conversations.

%% file: sections/VII-acknowledgement.tex
Students Ruoyu Hu and Neophytos Polydorou were supported by UK Research and Innovation [UKRI Centre for Doctoral Training in AI for Healthcare grant number EP/S023283/1]. We appreciate the general support from the Empowered Human Foundation in Canada and the UK.